\documentclass[letterpaper, 10 pt, conference]{ieeeconf}

\IEEEoverridecommandlockouts
\usepackage{commands}
\usepackage{caption}
\usepackage{hyperref}

\title{\LARGE \bf 
VBGS-SLAM: Variational Bayesian Gaussian Splatting Simultaneous Localization and Mapping
}

\author{Anonymous Authors}

\author{Yuhan Zhu, Yanyu Zhang, Jie Xu, Wei Ren
\thanks{Y. Zhu, Y. Zhang, J. Xu, and W. Ren are with the Department of Electrical and Computer Engineering, University of California, Riverside, CA, 92521, USA. Email: \{yzhu275, yzhan831, jxu150, weiren\}@ucr.edu.}%
}

\begin{document}

\maketitle
\thispagestyle{empty}
\pagestyle{empty}

\begin{abstract}
3D Gaussian Splatting (3DGS) has shown promising results for 3D scene modeling using mixtures of Gaussians, yet its existing simultaneous localization and mapping (SLAM) variants typically rely on direct, deterministic pose optimization against the splat map, making them sensitive to initialization and susceptible to catastrophic forgetting as map evolves. We propose Variational Bayesian Gaussian Splatting SLAM (VBGS-SLAM), a novel framework that couples the splat map refinement and camera pose tracking in a generative probabilistic form. By leveraging conjugate properties of multivariate Gaussians and variational inference, our method admits efficient closed-form updates and explicitly maintains posterior uncertainty over both poses and scene parameters. This uncertainty-aware method mitigates drift and enhances robustness in challenging conditions, while preserving the efficiency and rendering quality of existing 3DGS. Our experiments demonstrate superior tracking performance and robustness in long sequence prediction, alongside efficient, high-quality novel view synthesis across diverse synthetic and real-world scenes.
\end{abstract}

\section{INTRODUCTION}
Simultaneous localization and mapping (SLAM) is a cornerstone of modern computer vision, enabling applications in robotics, autonomous driving, and 3D reconstruction. Conventional SLAM systems typically model the environment with either sparse point clouds or dense voxels \cite{niessner2013real, Zhang2019eit, dai2017bundlefusion, newcombe2011kinectfusion, zhang2025iros}. While effective, these discrete representations are memory-demanding and struggle to fill in unobserved regions, leading to incomplete reconstructions \cite{zhu2022nice}. Implicit geometry representation using neural networks such as Multi-Layer Perceptron (MLP) has emerged recently as a promising approach, as they offer compact representation \cite{mescheder2019occupancy}, robustness to noise and errors \cite{huang2021di}, and flexibility in resolution extraction \cite{park2019deepsdf}. Implicit geometry representations offer compact and flexible scene modeling with MLP. However, a single global MLP must be retrained whenever new observations arrive \cite{zhang2025nerf}, leading to prohibitive training and inference times on large-scale scenes (e.g., office buildings or city blocks). Therefore, several works \cite{mescheder2019occupancy, park2019deepsdf, yang2022vox} have introduced scaling data to voxels as an alternative representation of 3D scenes, achieving orders of magnitude faster training time, while still preserving the power of compact, robust, and flexible representation of implicit mapping. However, these neural implicit representations require expensive per-pixel raycasting to render \cite{matsuki2024gaussian}, and map updates can overwrite prior content, leading to catastrophic forgetting. 

Recently, 3DGS has demonstrated the effectiveness of 3D scene representation, combining the practical benefits of point map with the differentiability and fidelity of neural scene representation. New evidence can edit and prune local splats without global retraining, making 3DGS particularly adaptable for streaming updates in SLAM. Consequently, 3DGS has been adopted in recent dense visual SLAM systems such as SplaTAM \cite{keetha2024splatam} and MonoGS \cite{matsuki2024gaussian}, outperforming earlier neural-implicit solutions. Despite their success, GS-based SLAM systems suffer from fragile initialization and tightly coupled, computation-heavy optimization, which limits scalability to long sequences. Moreover, deterministic gradient-based optimization provides no principled mechanism for uncertainty modeling, making these systems sensitive to initialization and prone to catastrophic forgetting \cite{french1999catastrophic}.

A recent alternative, Variational Bayesian Gaussian Splatting (VBGS) \cite{van2024variational}, frames 3D scene representation learning as a variational inference problem over model parameters, enabling closed-form updates that efficiently incorporate sequential observations and naturally quantify uncertainty. However, VBGS assumes known camera poses and data statistics, confining it to offline reconstruction. In the SLAM setting, pose and map uncertainties are tightly coupled and must be inferred jointly online, which prevents its direct application. 

To address these limitations, we introduce \textbf{V}ariational \textbf{B}ayesian \textbf{G}aussian \textbf{S}platting \textbf{S}imultaneous \textbf{L}ocalization \textbf{a}nd \textbf{M}apping (VBGS-SLAM), a fully probabilistic RGB-D SLAM framework that combines Gaussian Splatting with Variational Bayesian Inference:

\begin{itemize}
    \item We formulate a generative mixture model comprising 3-D Gaussians and SE(3) poses expressed in Lie groups, and derive closed-form variational update rules for both the splat map and the camera pose.
    \item Our variational formulation couples the Gaussian map and SE(3) pose variables, allowing pose uncertainty to be carried through the mapping step rather than treated post-hoc. This tighter probabilistic coupling produces more accurate maps that stabilize and improve pose tracking. Moreover, because of the closed-form variational updates, VBGS-SLAM updates map parameters and poses far more efficiently than gradient-based Gaussian Splatting pipelines.
    \item We benchmark on Replica, TUM-RGBD, and AR-TABLE datasets, demonstrating state-of-the-art accuracy and runtime efficiency. 
\end{itemize}

\section{RELATED WORKS}
\label{sec:related works}

\subsection{Traditional Visual SLAM}
Among the traditional visual SLAM literature, existing approaches can be broadly classified into two categories: sparse SLAM \cite{Mur2015, Mur2017, engel2014lsd, Boretti2022, Zhang2023} and dense SLAM \cite{newcombe2011kinectfusion, dai2017bundlefusion, murai2025mast3r}. Early systems such as MSCKF \cite{Mourikis2007} and ORB-SLAM \cite{Mur2015} relied on sparse point features to establish correspondences across frames, while LSD-SLAM \cite{engel2014lsd} proposed a direct point tracking approach based on the epipolar geometry. These methods primarily targeted localization accuracy, but provided limited scene information for a detailed 3D representation. In contrast, dense visual SLAM emphasizes generating rich 3D maps. KinectFusion \cite{newcombe2011kinectfusion} demonstrated the feasibility of real-time dense reconstruction using RGB-D cameras to estimate volumetric surface models. Subsequent research extended these foundations to improve robustness against fast motion, dynamic environments, and large-scale scenes. For instance, BundleFusion \cite{dai2017bundlefusion} introduced a globally consistent room reconstruction by combining dense frame tracking within a sliding window. More recently, MASt3R-SLAM \cite{murai2025mast3r} incorporated two-view 3D priors to strengthen real-time monocular dense SLAM without requiring assumptions on camera models. Despite these advances, achieving high-fidelity, real-time dense reconstruction in unconstrained environments remains an open challenge.

\subsection{Learning-based Visual SLAM}
With the rapid advancement of GPU computing, the SLAM community has increasingly shifted toward learning-based approaches in recent years, where the environment is implicitly encoded within neural networks. A milestone in this direction is the Neural Radiance Fields (NeRF) \cite{mildenhall2021nerf}, which employs a fully connected neural network to represent a scene. Building on this foundation, works such as NeRF-VINS \cite{Katragadda2024} and NeRF-VIO \cite{zhang2025nerf} leveraged pretrained NeRF models as prior maps and integrated them into SLAM architectures. However, the inherently non-differentiable nature of NeRF prevents these systems from supporting online map updates, limiting their applicability in real-time SLAM systems.

More recently, 3D Gaussian Splatting (3DGS) \cite{kerbl20233d} has emerged as a powerful alternative for scene representation. In this paradigm, the environment is modeled as a collection of 3D Gaussian ellipsoids. This explicit yet differentiable representation allows for rendering at real-time rates, directly addressing the speed limitations of NeRF. One of the pioneering works is Gaussian Splatting SLAM \cite{matsuki2024gaussian}, which first demonstrated the feasibility of employing 3DGS in monocular SLAM and rendering within a single Gaussian-based representation. Subsequent works have extended to address dynamic environments \cite{kong2024dgs, li2025dynagslam}, scalability to large-scale scenes \cite{xin2025large}, and multi-sensor integration \cite{hong2025gs}. Collectively, these efforts highlight the promise of 3DGS as a versatile and efficient scene representation for SLAM, striking a balance among rendering quality, real-time performance, and structural fidelity across diverse settings.

\section{PRELIMINARIES}\label{sec:Preliminaries}

\subsection{3D Gaussian Splatting}

3D Gaussian Splatting (3DGS) \cite{kerbl20233d} represent a scene as a collection of Gaussian ellipsoids $\mathcal{G}$. Each Gaussian $\mathcal{G}_k$, for $k = 1, \cdots, K$, is characterized by its position and ellipsoidal shape, defined by a mean $\boldsymbol{\mu}^{\mathcal{W}}_k$ and covariance $\mathbf{\Sigma}^{\mathcal{W}}_k$ in the world frame $\{\mathcal{W}\}$. In addition, each $\mathcal{G}_k$ encodes optical attributes including color $\mathbf{c}_k$ and opacity $\mathbf{\alpha}_k$. Then, the Gaussian is transformed from the world frame to the camera frame $\{\mathcal{C}_t\}$ with the known camera pose ${}^{\mathcal{C}_t}\mathbf{T}_{\mathcal{W}}$. For brevity, we denote this relative transformation as $\mathbf{T}_t$ in subsequent sections. Finally, images are synthesized by projecting and alpha-compositing these Gaussians on the image plane, and parameters are optimized by gradient descent \cite{kingma2014adam} to minimize the discrepancy between rendered and observed images.

\subsection{Variational Bayesian Gaussian Splatting}

Variational Bayesian Gaussian Splatting (VBGS) \cite{van2024variational} frames 3D scene representation learning as a variational inference problem over model parameters. In this probabilistic framework, the scene is modeled as a mixture of $K$ Gaussian components. Each component $\mathcal{G}_k$ is represented by two conditionally independent modalities: the spatial position $\mathbf{s} \sim \mathcal{N}(\boldsymbol{\mu}_{k,s}, \mathbf{\Sigma}_{k,s})$ and the color $\mathbf{c} \sim\mathcal{N}(\boldsymbol{\mu}_{k,c}, \mathbf{\Sigma}_{k,c})$, which are both modeled as multivariate Normal (MVN) distributions \cite{tong2012multivariate}. The full generative model can be represented by a Bayesian network and factorized as:
\begin{equation}
    \begin{aligned}
        &p(\mathbf{s}, \mathbf{c}, \mathbf{z}, \boldsymbol{\mu}_s, \mathbf{\Sigma}_s, \boldsymbol{\mu}_c, \mathbf{\Sigma}_c, \boldsymbol{\pi}) \\
        =&\Big(\prod_{n=1}^{N} p(\mathbf{s}_n | z_n, \boldsymbol{\mu}_s, \mathbf{\Sigma}_s) p(\mathbf{c}_n | z_n, \boldsymbol{\mu}_c, \mathbf{\Sigma}_c) p(z_n | \boldsymbol{\pi}) \Big)\\
        &\Big( \prod_{k=1}^{K} p(\boldsymbol{\mu}_{k,s}, \mathbf{\Sigma}_{k,s}) p(\boldsymbol{\mu}_{k,c}, \mathbf{\Sigma}_{k,c}) \Big)p(\boldsymbol{\pi}),
    \label{eq:vbgs_prior}
    \end{aligned}
\end{equation}
where $\mathbf{z} = [z_1, \cdots, z_n, \cdots, z_N]^\top$ represents the point assignment, modeled by a categorial distribution \cite{agresti2005bayesian} parameterized by the weights $\boldsymbol{\pi}$, and $n$ denotes the number of 3D point in one training step, for $n = 1, \cdots, N$. The parameters defined in Eq.~\eqref{eq:vbgs_prior} are treated as latent random variables, allowing refinement via variational inference utilizing the properties of conjugate priors. Namely, VBGS uses Normal Inverse Wishart (NIW) \cite{nydick2012wishart} to parameterize the mean and covariance of an MVN, and uses the Dirichlet distribution \cite{ng2011dirichlet} to parameterize $\boldsymbol{\pi}$. 


Since the true posterior is intractable, VBGS uses a mean-field approximation, which assumes variational posterior factorizes across the latent variables. Specifically, the variational posterior distribution $q$ is decomposed as: 
\begin{equation}
    \begin{aligned}
        &q(\mathbf{z}, \boldsymbol{\mu}_s, \mathbf{\Sigma}_s, \boldsymbol{\mu}_c, \mathbf{\Sigma}_c, \boldsymbol{\pi}) = \\
        &\Big( \prod_{n=1}^{N} q(z_n) \Big)
        \Big( \prod_{k=1}^{K} q(\boldsymbol{\mu}_{k,c}, \mathbf{\Sigma}_{k,c}) \Big)
        \Big( \prod_{k=1}^{K} q(\boldsymbol{\mu}_{k,s}, \mathbf{\Sigma}_{k,s}) \Big)
        q(\boldsymbol{\pi}),
    \label{eq:vbgs_posterior}
    \end{aligned}
\end{equation}
where the approximate posteriors are selected from the same family as their corresponding priors.

Then, the parameters are estimated by minimizing the Kullback-Leibler (KL) divergence \cite{joyce2011kullback} between the approximate posterior and the true posterior as: 
\vspace{-5pt}
\begin{equation}
    \begin{aligned}
        \argmin_{\boldsymbol{\lambda}} \mathbb{D}_{\text{KL}} &\left[ q(\boldsymbol{\theta}) \, || \, p( \boldsymbol{\theta}| \mathbf{s}, \mathbf{c}) \right],
     \label{eq:vbgs_kld}
    \end{aligned}
\end{equation}
where $\boldsymbol{\lambda}$ denotes the variational parameters of the latent variable $\boldsymbol{\theta} = \{\mathbf{z},\boldsymbol{\mu}_s,\boldsymbol{\Sigma}_s,\boldsymbol{\mu}_c,\boldsymbol{\Sigma}_c,\boldsymbol{\pi}\}$ in both prior and posterior distribution. 

VBGS then relies on coordinate ascent variational inference (CAVI) \cite{blei2017variational} and the conjugate properties of the exponential distribution family to derive a closed-form update rule for each variational parameter in $\boldsymbol{\lambda}$. For a more comprehensive description, readers are referred to \cite{van2024variational}.
\begin{figure*}[t]
    \centering
    \includegraphics[width=1.0\linewidth]{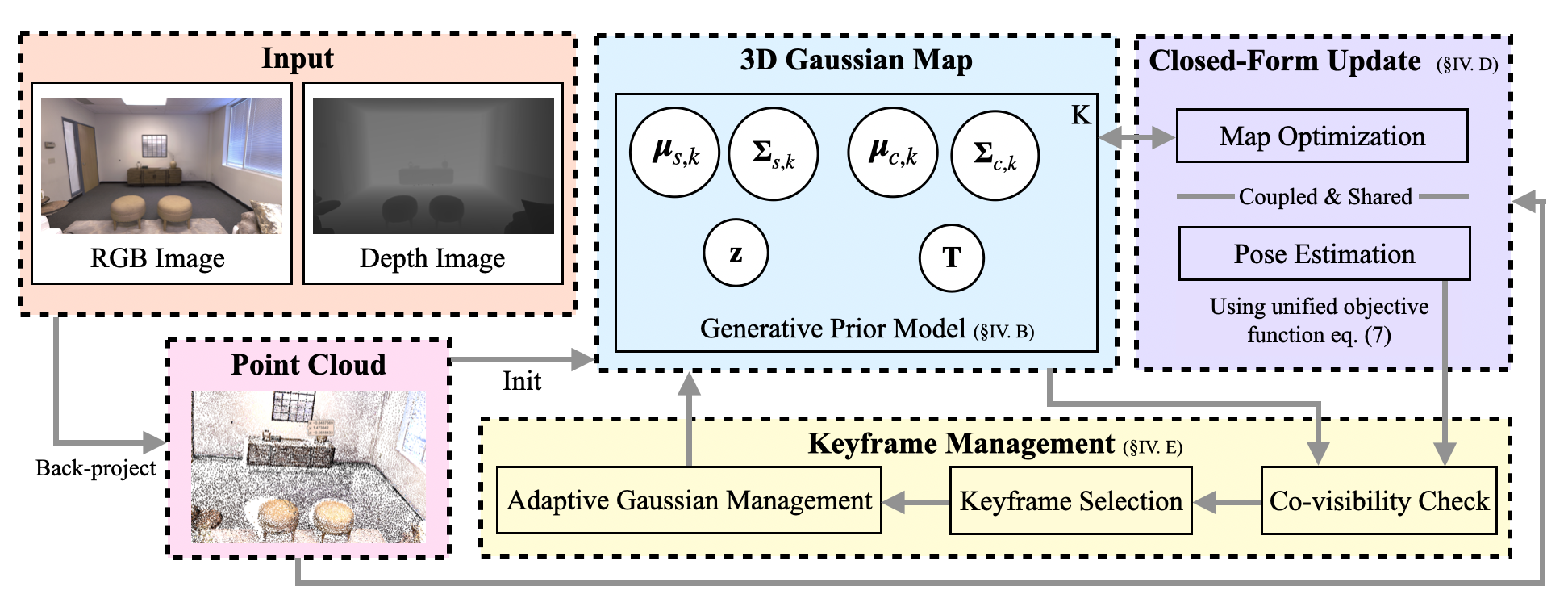}
    \captionsetup{aboveskip=2pt,belowskip=0pt}
    \caption{VBGS-SLAM System Overview: Our method takes RGB-D images as input and initializes a point cloud via back-projection. This point cloud seeds a probabilistic 3D Gaussian map parameterized by spatial and color distribution under a generative prior model. The pipeline can be interpreted as a closed-form variational inference framework that jointly optimizes the Gaussian map and estimates the camera pose through a unified objective in Eq.~\eqref{eq:vbgs_slam_kld_fixed}. The online processing integrates new RGB-D image pairs, where keyframe management governs Gaussian adaption, keyframe selection, and co-visibility checking. This unified strategy enables efficient real-time SLAM by coupling mapping and tracking within a shared closed-form update loop. }
    \label{fig:vbgs}
\end{figure*}
\vspace{-5pt}
\section{VBGS-SLAM}\label{sec:vbgs_slam}
The goal of VBGS-SLAM is to estimate the camera pose $\mathbf{T}_t$, while concurrently constructing a map encoded as a set of Gaussian ellipsoids $\mathcal{G}$. As shown in Fig.~\ref{fig:vbgs}, the system is initialized by back-projecting the synchronized RGB-D image pairs into a dense point cloud. The point cloud is then used to update the generative model via a closed-form probabilistic framework. To maintain both compactness and accuracy of the map, the framework employs keyframe selection and adaptive Gaussian management, enabling dynamic insertion and pruning of Gaussians to regulate map density, thereby avoiding the computational overhead of gradient-based optimization methods.
\vspace{-5pt}

\subsection{System Initialization}
The system is initialized with the camera intrinsic matrix $\mathcal{K}$, the initial RGB-D image pair $(C_0, D_0)$, and the first camera pose $\mathbf{T}_0$. A dense 3D point cloud is first generated by back-projecting image pixels into 3D space using their depth values and transforming the resulting points into the world frame according to the initial pose. To mitigate the sensor noise and reduce redundancy, this raw point cloud is processed through a voxel grid downsampling filter. From this filtered set, we construct the initial 3D Gaussian map. Each point seeds a single Gaussian primitive, where the mean $\boldsymbol{\mu}_{k, s}$ is set to the point's 3D coordinates. An anisotropic covariance matrix $\mathbf{\Sigma}_{k, s}$, is derived from the spatial distribution of the neighboring points.
\vspace{-10pt}
\subsection{Generative Prior Model} \label{sec:Generative model}
To incorporate VBGS into a real-time SLAM framework, we formulate a unified probabilistic framework that jointly performs pose tracking and dense mapping. By modeling the camera pose as a latent variable, our method enables the system to capture the uncertainty in both scene geometry and camera motion. Specifically, we explicitly model the spatial position of each 3D point generated from an observed RGB-D image pair, which is conditioned on the camera pose $\mathbf{T}_t$. The full generative model is defined as:
\begin{equation}
    \begin{aligned}
        &p(\mathbf{s}, \mathbf{c}, \mathbf{z}, \boldsymbol{\mu}_s, \mathbf{\Sigma}_s, \boldsymbol{\mu}_c, \mathbf{\Sigma}_c, \boldsymbol{\pi}, \mathbf{T}_t) \\
        =&\Big(\prod_{n=1}^{N} p(\mathbf{s}_n | z_n, \boldsymbol{\mu}_s, \mathbf{\Sigma}_s, \mathbf{T}_t) p(\mathbf{c}_n | z_n, \boldsymbol{\mu}_c, \mathbf{\Sigma}_c) p(z_n | \boldsymbol{\pi}) \Big)\\
        &\Big( \prod_{k=1}^{K} p(\boldsymbol{\mu}_{k,s}, \mathbf{\Sigma}_{k,s}) p(\boldsymbol{\mu}_{k,c}, \mathbf{\Sigma}_{k,c}) \Big)p(\mathbf{T}_t|\boldsymbol{\mu}_{\xi, t}, \boldsymbol{\Sigma}_{\xi, t})p(\boldsymbol{\pi}),
    \label{eq:vbgs_slam_prior}
    \end{aligned}
\end{equation}
where $(\mathbf{s}_n, \mathbf{c}_n)$ are obtained by back-projecting RGB-D image pairs. 

Since the back-projection defines the spatial component $\mathbf{s}_n$ in the camera frame, the spatial term of the Gaussian mixture model is reformulated to incorporate the camera pose $\mathbf{T}_t$ as:
\begin{equation}
p(\mathbf{s}_n \mid z_n{=}k, \mathbf{T}_t)
= \mathcal{N}\!\big(\mathbf{T}_t \odot \boldsymbol{\mu}_{k,s},\; \mathbf{R}_t\mathbf{\Sigma}_{k,s}\mathbf{R}_t^\top),
\end{equation}
where the frame transformation $\odot$ is defined by
$
\mathbf{T}_t \odot \boldsymbol{\mu} \triangleq \mathbf{R}_t \boldsymbol{\mu} + \mathbf{t}_t,
$
with $\mathbf{T}_t=\begin{bmatrix}\mathbf{R}_t & \mathbf{t}_t\\ \mathbf{0}^\top & 1\end{bmatrix}\in SE(3)$. This formulation directly links the spatial distribution of the Gaussian ellipsoids to the current camera pose, thereby coupling the mapping and tracking process. The camera pose $\mathbf{T}_t$ is modeled as an MVN distribution $p(\mathbf{T}_t) \sim \mathcal{N}(\boldsymbol{\xi}|\boldsymbol{\mu}_{\xi, t}, \boldsymbol{\Sigma}_{\xi, t})$ in the $\mathfrak{se}(3)$ space, with \(\mathbf{T}_t\) obtained via the exponential map, which is standard due to the locally Euclidean structure of $\mathfrak{se}(3)$ and its compatibility with linearization.

\subsection{Variational Posterior}
By treating the camera pose as a latent variable within the generative model, its distribution is explicitly included in the variational posterior, enabling the joint inference of both pose and map parameters. To make the inference tractable, we employ a mean-field approximation that factorizes the joint posterior as:
\begin{equation}
    \begin{aligned}
        &q(\mathbf{z}, \boldsymbol{\mu}_s, \mathbf{\Sigma}_s, \boldsymbol{\mu}_c, \mathbf{\Sigma}_c, \boldsymbol{\pi}, \mathbf{T}_t)
        =\Big( \prod_{n=1}^{N} q(z_n) \Big) \\
        &\Big( \prod_{k=1}^{K} q(\boldsymbol{\mu}_{k,c}, \mathbf{\Sigma}_{k,c}) \Big)
        \Big( \prod_{k=1}^{K} q(\boldsymbol{\mu}_{k,s}, \mathbf{\Sigma}_{k,s}) \Big) 
        q(\mathbf{T}_t)q(\boldsymbol{\pi}),
    \label{eq:vbgs_slam_posterior}
    \end{aligned}
\end{equation}
where the pose factor is parameterized on the Lie group as $\boldsymbol{\xi}_t \sim \mathcal{N}(\boldsymbol{\mu}_{\xi,t}, \boldsymbol{\Sigma}_{\xi,t})$. Variational inference seeks the member of the exponential family that best approximates the true posterior. Concretely, we minimize the KL divergence between \(q(\cdot)\) and \(p(\cdot \mid \mathbf{s},\mathbf{c})\), which is equivalent to maximizing the evidence lower bound (ELBO) \cite{jordan1999introduction}, thus tying posterior approximation to data fit as:
\begin{equation}
\begin{aligned}
    &\mathbb{D}_{\mathrm{KL}}\!\big[q(\boldsymbol{\theta}) \,\|\, p(\boldsymbol{\theta}\mid \mathbf{s},\mathbf{c})\big]\\
&= \log p(\mathbf{s},\mathbf{c}) - \underbrace{\Big(\mathbb{E}_{q}\![\log p(\mathbf{s},\mathbf{c},\boldsymbol{\theta})] - \mathbb{E}_{q}\![\log q(\boldsymbol{\theta})]\Big)}_{\mathcal{L}(q)\ \text{(ELBO)}},
\end{aligned}
\end{equation}
with \(\boldsymbol{\theta}\!=\!\{\mathbf{z},\boldsymbol{\mu}_s,\boldsymbol{\Sigma}_s,\boldsymbol{\mu}_c,\boldsymbol{\Sigma}_c,\boldsymbol{\pi},\mathbf{T}_t\}\).
Hence, minimizing the KL finds the closest joint posterior over pose and map while maximizing a lower bound on the log evidence. We therefore optimize the variational parameters $\boldsymbol{\lambda}$, which parameterize each latent variable in $\boldsymbol{\theta}$ defined above:
\vspace{-10pt}
\begin{equation}
\arg\min_{\boldsymbol{\lambda}}\ 
\mathbb{D}_{\mathrm{KL}}\!\left[
q_{\boldsymbol{\lambda}}(\boldsymbol{\theta})
\;\big\|\;
p(\boldsymbol{\theta}\mid \mathbf{s}, \mathbf{c})
\right]
\;\;\Longleftrightarrow\;\;
\arg\max_{\boldsymbol{\lambda}}\ \mathcal{L}\!\left(q_{\boldsymbol{\lambda}}\right).
\label{eq:vbgs_slam_kld_fixed}
\end{equation}

By comparing the base prior in Eq.~\eqref{eq:vbgs_prior} with its SLAM counterpart in Eq.~\eqref{eq:vbgs_slam_prior}, the posterior factorization in Eq.~\eqref{eq:vbgs_posterior} and in Eq.~\eqref{eq:vbgs_slam_posterior}, and the objective in Eq.~\eqref{eq:vbgs_kld} and in Eq.~\eqref{eq:vbgs_slam_kld_fixed}, we explicitly promote the camera pose $\mathbf{T}_t$ from a conditioned quantity to a latent variable endowed with its own prior and variational factor. Then, we minimize the KL divergence to jointly infer the pose trajectory and the Gaussian map parameters, along with assignments $\mathbf{z}$ and mixture weights $\boldsymbol{\pi}$. At this stage, the problem transitions from 3D reconstruction with known poses to a full SLAM formulation in which mapping and tracking are coupled and estimated simultaneously.
\vspace{-10pt}
\subsection{Closed-Form Update}

VBGS \cite{van2024variational} uses mean-field variational inference with coordinate-ascent updates under conjugate priors. This yields closed-form updates of component posteriors and point assignment. However, these updates assume known poses. To couple mapping and tracking, we promote the camera pose to a latent variable and take expectations over its variational factor $q(\mathbf{T}_t)$ inside the point assignment update. This preserves closed-form structure while propagating pose uncertainty into assignments and spatial statistics. By gathering terms involving $z_n$ and taking the derivative with respect to $q(z_n)$, the soft assignment for point $n$ to component $k$ is:
\vspace{-5pt}
\begin{equation}
\begin{aligned}
    &\log \gamma_{nk}\\
    &= \mathbb{E}_{q(\boldsymbol{\mu}_{k,s}, \mathbf{\Sigma}_{k,s})q(\mathbf{T}_t)} \left[ \log p(\mathbf{s}_n | z_n, \mathbf{T}_t \odot \boldsymbol{\mu}_{k,s},\; \mathbf{R}_t\mathbf{\Sigma}_{k,s}\mathbf{R}_t^\top \right] \\
    &+ \mathbb{E}_{q(\boldsymbol{\mu}_{k,c}, \mathbf{\Sigma}_{k,c})} \left[ \log p(\mathbf{c}_n | \boldsymbol{\mu_{k,c}}, \mathbf{\Sigma}_{k,c}) \right] \\
    &+ \mathbb{E}_{q(\boldsymbol{\pi})} \left[ \log p(\boldsymbol{\pi}) \right] - \log Z_n, 
\end{aligned}
\end{equation}
where \(\mathbb{E}_q[\,\cdot\,]\) denotes expectation with respect to the specified variational factors. The first term in this equation, which represents the expected log-likelihood for the spatial position, demonstrates that a 3D data point's assignment to a specific Gaussian component is now explicitly coupled with the probabilistic estimate of the camera pose. The remaining terms are unaffected, as color and mixture weight assignments are independent of pose. This ensures point assignments remain robust to camera pose uncertainty while preserving the independence of the color model.

The inclusion of the latent camera pose $\mathbf{T}_t$ fundamentally alters the update rule for spatial properties of our Gaussian map. Our approach propagates pose uncertainty into the map by basing updates on an expectation over the pose distribution $q(\mathbf{T}_t)$. As a result, the sufficient statistics for the spatial posterior $\mathcal{T}(\mathbf{s}_n)$ are now dependent on the pose posterior, defined as:
\begin{equation}
\mathcal{T}(\mathbf{s}_n)
\;\triangleq\;
\big(\,\mathbf{T}_t^{-1} \odot \mathbf{s}_n,\;\; \mathbf{R}_t^\top\, \mathbf{s}_n \mathbf{s}_n^\top\, \mathbf{R}_t\,\big).
\end{equation}
Crucially, our method introduces a probabilistic framework for camera pose estimation. We formulate the pose estimation problem as a probabilistic update, rather than a direct optimization using gradient descent as seen in methods like MonoGS \cite{matsuki2024gaussian}. 

To derive a closed-form solution for the approximate posterior $q(\mathbf{T}_t)$, we address the non-linear relationship between the pose $\mathbf{T}_t$ and the Gaussian map $\mathcal{G}$. This is achieved by linearizing the spatial likelihood around the current pose estimate $\boldsymbol{\mu}_{\xi, t}$ using first-order Taylor expansion on the tangent space:
\begin{equation}
    \mathbf{T}_t\odot \boldsymbol{\mu}_{k,s} \approx \bar{\mathbf{T}}_t\odot \boldsymbol{\mu}_{k,s} + \mathbf{G}_{k, s}\delta \boldsymbol{\xi},
\end{equation}
where $\mathbf{T}_t \;=\; \bar{\mathbf{T}}_t\,\exp\!\big([\delta\boldsymbol{\xi}]^\wedge\big)$, and $\bar{\mathbf{T}}_t=\exp\!\big([\boldsymbol{\mu}_{\xi,t}]^\wedge\big)$. And  $[\,\cdot]\,^\wedge$ denotes the wedge operator that maps a 6-vector $\xi$ to its corresponding $4\times 4$ matrix.The corresponding Jacobian of the transformed mean with respect to $\boldsymbol{\xi}$ is computed as:
\begin{equation}
    \mathbf{G}_{k,s}=\frac{\partial \mathbf{T}_t \odot\boldsymbol{\mu}_{k,s}}{\partial \boldsymbol{\xi}} \bigg|_{\boldsymbol{\xi}=\boldsymbol{\mu}_{\xi, t}} = \big[\,\bar{\mathbf{R}}_t\;\; -\,\bar{\mathbf{R}}_t[\boldsymbol{\mu}_{k,s}]_\times\,\big]\ ,
\end{equation}
where $[\,\cdot]\,_\times$ is the skew operator.
For the covariance action, for brevity, we define the covariance at mean pose:
\begin{equation}
\bar{\mathbf{R}}_t\,\boldsymbol{\Sigma}_{k,s}\,\bar{\mathbf{R}}_t^\top
\ \triangleq\ \tilde{\boldsymbol{\Sigma}}_{k,s}.
\end{equation}
This linearization allows us to derive a set of closed-form update rules for the camera pose posterior:
\begin{align}
\mathbf{\Sigma}_{\xi,t+1}^{-1} 
&= \sum_{n=1}^N\sum_{k=1}^K
   \gamma_{nk}\mathbf{G}_{k,s}^\top\tilde{\boldsymbol{\Sigma}}_{k,s}^{-1}\mathbf{G}_{k,s}
   + \mathbf{\Sigma}_{\xi,t}^{-1}, \label{eq:sigma_update} \\[6pt]
\boldsymbol{\mu}_{\xi,t+1} 
&= \mathbf{\Sigma}_{\xi,t+1}\Big(
   \sum_{n=1}^N\sum_{k=1}^K 
   \gamma_{nk}\mathbf{G}_{k,s}^\top\tilde{\mathbf{\Sigma}}_{k,s}^{-1}
   (\mathbf{s}_n - \bar{\mathbf{T}}_t \odot \boldsymbol{\mu}_{k,s}) \notag \\
&\qquad\qquad\quad 
   + \mathbf{\Sigma}_{\xi,t}^{-1}\boldsymbol{\mu}_{\xi,t}\Big). 
   \label{eq:mu_update}
\end{align}
These updates for both the map and the pose can be computed concurrently using a single pass to calculate the sufficient statistic. This unified approach largely reduces the computational time required for differential rendering and separate pose optimization, making the system suitable for real-time SLAM applications. 

\subsection{Keyframe Management}
To maintain a compact and high-resolution map, we introduce two complementary strategies: \textit{Keyframe Selection} and \textit{Adaptive Gaussian Management}.

\subsubsection{Keyframe Selection}
Inspired by MonoGS \cite{matsuki2024gaussian}, we propose a dual-criteria strategy for keyframe selection that combines Gaussian co-visibility with a temporal constraint. The co-visibility criterion triggers a new keyframe only when the current image observes sufficiently unseen Gaussians. This prevents the insertion of redundant keyframes for already well-represented regions. On the other hand, the temporal constraint enforces a maximum frame interval between successive keyframes. This prevents the system from going too long without inserting a keyframe, which could otherwise lead to accumulated drift in scenarios involving slow camera motion. Together, these criteria balance redundancy reduction with trajectory coverage. 

Selected keyframes are stored in a fixed-size keyframe buffer and optimized using a sliding-window variational inference scheme. At each optimization step, the approximate posterior distributions over the poses of all keyframes in the buffer are jointly optimized together with the associated Gaussian map parameters, while older keyframes are marginalized. This joint optimization allows pose uncertainty to propagate consistently across multiple frames while maintaining computational tractability.

When a new keyframe is selected, our system generates a new set of Gaussian components by back-projecting the current RGB-D image pair. In contrast to MonoGS, we apply an additional filter to this point cloud to avoid redundant insertion of Gaussians in regions that are already well-represented. This filter reduces the number of parameters that must be maintained during mapping, thereby improving computational efficiency in long-sequence SLAM and providing robustness in loop-closure detection.

\subsubsection{Adaptive Gaussian Management}
We further employ an adaptive mechanism to refine the Gaussian map, which addresses two complementary challenges: \textit{insufficient coverage} and \textit{excessive redundancy}. 

In regions with insufficient coverage, where only a few Gaussians can be observed, new Gaussians are introduced to fill the gaps and ensure adequate representation. Apart from normal Gaussian Splatting where the two types of regions can be filtered with masking component with a large gradient, we propose to use a soft Manhattan distance-based analysis to determine the unconstructed region and insert new components initiated from the current RGB-D image pair. In addition, we applied a different prune strategy, where the components with unchanged prior $\pi$ are re-spawned. This ensures that the Gaussian set remains representative of the underlying data distribution, while avoiding redundant cloning and capturing fine-grained geometric detail more robustly.

\section{EXPERIMENTS}


In this section, we compare the accuracy and efficiency of localization and 3D reconstruction between five open-source methods and our proposed approach, using both real and synthetic datasets. We first introduce the experiment setup in Sec.~\ref{sec:exp_setup}. Then, we evaluate each method in terms of localization accuracy, map reconstruction quality, and rendering time across different scenarios in Sec.~\ref{sec:loc_performance} and Sec.~\ref{sec:recon_performance}. All experiments are performed on a desktop equipped with an Intel Core i9-12900K processor and a single NVIDIA GeForce RTX 4090 GPU. In addition, we conduct an ablation study to assess the effect of integrating inertial measurement unit (IMU) data on tracking accuracy and rendering performance in Sec.~\ref{sec:ablation_study}.

\subsection{Experiment Setup}
\label{sec:exp_setup}

\subsubsection{Datasets}
To evaluate the performance of our proposed VBGS-SLAM and other baselines, we conducted experiments on a diverse set of datasets, encompassing both synthetic and real-world environments with varying levels of complexity and sensor modalities. Specifically, we evaluate using (1) the synthetic Replica Dataset \cite{replica19arxiv}, (2) the real-world TUM-RGBD dataset \cite{sturm12iros}, and (3) the real-world AR-TABLE dataset \cite{Chen2023ICRA}. Detailed comparison of these datasets is provided in Table \ref{tab:dataset_summary}. 

\begin{table}[h]
	\centering
	\caption{Summary of dataset characteristics. AR-TABLE dataset contains loop trajectories.}
	\begin{tabular}{ c | c c c }
		\toprule
		\textbf{Dataset} & \textbf{Type} & \textbf{Sensor} & \textbf{Motion Blur}\\
		\midrule
		Replica~\cite{replica19arxiv} & Synthetic & RGB-D & $\times$ \\
        TUM-RGBD~\cite{sturm12iros} & Real-world & RGB-D \& Acc & $\checkmark$\\
        AR-Table~\cite{Chen2023ICRA} & Real-world & RGB-D \& IMU & $\checkmark$ \\
		\bottomrule
	\end{tabular}
	\label{tab:dataset_summary}
\end{table}





\subsubsection{Metrics}
To evaluate the performance of VBGS-SLAM and the baseline methods, we use absolute trajectory error (ATE) \cite{Zhang2018} to measure localization accuracy, and PSNR, SSIM, LPIPS, and FPS \cite{sandstrom2023point} to assess 3D reconstruction quality and efficiency.

\subsubsection{Baselines}

We compare our VBGS-SLAM against \textit{state-of-the-art} dense visual SLAM approaches, including NICE-SLAM \cite{zhu2022nice}, Vox-Fusion \cite{yang2022vox}, Point-SLAM \cite{sandstrom2023point}, SplaTAM \cite{keetha2024splatam}, and MonoGS \cite{matsuki2024gaussian} as they represent the current state of the art across the principal designs for dense visual SLAM: grid/voxel neural fields, point-based neural mapping, and 3D Gaussian Splats. Among them, we highlight MonoGS and SplaTAM as our primary points of comparison, since they also represent scenes with 3D Gaussian splats and are therefore most directly aligned with our approach. A summary of those baselines is provided in Table~\ref{tab:baseline_summary}.
\setlength{\tabcolsep}{3.5pt}
\begin{table}[t]
	\centering
	\caption{Baselines comparison.}
	\begin{tabular}{ c | c c c}
		\toprule
		\textbf{Method} & \textbf{Scene repr.} & \textbf{Input} & \textbf{Limitations}\\
		\midrule
		NICE-SLAM~\cite{zhu2022nice} & Implicit SDF & RGB-D & GPU\text{-}intensive\\
        Vox-Fusion~\cite{yang2022vox} & Implicit Voxel TSDF & RGB-D & Resolution \\
        Point-SLAM~\cite{sandstrom2023point} & Point NeRF & RGB-(D) & Sensitive to init \\
        SplaTAM~\cite{keetha2024splatam} & 3D Gaussian splats & RGB-(D) & Opacity tuning \\
        MonoGS~\cite{matsuki2024gaussian} & 3D Gaussian splats & RGB-(D) & Sensitive to init\\
		\bottomrule
	\end{tabular}
	\label{tab:baseline_summary}
    \vspace*{-10pt}
\end{table}

\setlength{\tabcolsep}{3.5pt}
\begin{table}[h]
	\centering
	\caption{The ATE of the estimated camera poses (cm) for five baseline methods and our VBGS-SLAM across different sequences of the Replica dataset.}
	\begin{tabular}{c|cccccccc|c}
		\toprule
		\textbf{Method} & \textbf{r0} & \textbf{r1} & \textbf{r2} & \textbf{o0} & \textbf{o1} & \textbf{o2} & \textbf{o3} & \textbf{o4} & \textbf{Avg.}\\
		\midrule
		NICE-SLAM   & 0.97 & 1.31 & 1.07 & 0.88 & 1.00 & 1.06 & 1.10 & 1.13 & 1.07\\
        Vox-Fusion  & 1.37 & 4.70 & 1.47 & 8.48 & 2.04 & 2.58 & 1.11 & 2.94 & 3.09\\
        Point-SLAM  & 0.61 & 0.41 & 0.37 & \textbf{0.38} & 0.48 & 0.54 & 0.69 & 0.72 & 0.53\\
        MnonGS      & 0.47 & 0.43 & 0.31 & 0.78 & 0.57 & 0.31 & \underline{0.31} & 3.20 & 0.79\\
        SplaTAM     & \underline{0.31} & \underline{0.40} & 0.29 & 0.47 & \textbf{0.27} & \textbf{0.29} & 0.32 & \underline{0.55} & \underline{0.36}\\
    \midrule
    \textbf{Ours}   & \textbf{0.29} & \textbf{0.37} & \textbf{0.27} & \underline{0.39} & \textbf{0.27} & \underline{0.30} & \textbf{0.28} & \textbf{0.48} & \textbf{0.33}\\
		\bottomrule
	\end{tabular}
	\label{tab:replica_tracking}
    \vspace*{-8pt}
\end{table}

\begin{table}[h]
	\centering
	\caption{The ATE of the estimated camera poses (cm) for five baseline methods and our VBGS-SLAM across different sequences of the TUM-RGBD dataset.}
	\begin{tabular}{c|ccccc|c}
		\toprule
		\multirow{2}{*}{\textbf{Methods}} & \textbf{fr1/} & \textbf{fr1/} & \textbf{fr1/} & \textbf{fr2/} & \textbf{fr3/} & \multirow{2}{*}{\textbf{Avg.}}\\
        & \textbf{desk} & \textbf{desk2} & \textbf{room} & \textbf{xyz} & \textbf{off.} \\
		\midrule
		NICE-SLAM   & 4.26 & \underline{4.99} & 34.49 & 31.73 & \underline{3.87} & 15.87\\
        Vox-Fusion  & 3.52 & 6.00 & 19.53 & 1.49 & 26.01 & 11.31\\
        Point-SLAM  & 4.34 & \textbf{4.54} & 30.92 & 1.31 & \textbf{3.48} & 8.92\\
        MonoGS      & \textbf{3.35} & 6.54 & \textbf{11.86} & 1.34 & 5.41 & \textbf{5.70}\\
        SplaTAM     & 20.21 & - & - & 1.47 & - & -\\
    \midrule
    \textbf{Ours}   & \underline{3.49} & 6.54 & \underline{17.74} & \textbf{1.24} & 4.56 & \underline{7.69}\\
		\bottomrule
	\end{tabular}
	\label{tab:tum_tracking}
\end{table}
\vspace{-5pt}
\subsection{Localization Performance}
\label{sec:loc_performance}
The tracking performance on Replica, TUM-RGBD, and AR-TABLE datasets are presented in Table ~\ref{tab:replica_tracking}, ~\ref{tab:tum_tracking}, and ~\ref{tab:artable_tracking}. Across the three benchmarks, our method delivers the strongest average accuracy on Replica and AR-Table and ranks second on TUM-RGBD, while maintaining competitive results on sequence level. On Replica dataset, we achieved an average ATE of 0.33\,cm, improving upon SplaTAM with an average ATE of 0.36\,cm and Point-SLAM of 0.53\,cm by 8.3\% and 37.7\%, respectively, with best or near-best results across all sequences. On the real-world TUM-RGBD benchmark shown in Table~\ref{tab:tum_tracking}, our average ATE is 7.69\,cm, which trails SplaTAM’s 5.70 cm by 1.99 cm, but remains better than other baseline methods. The largest discrepancy appears on fr1/room, where a moving pedestrian induces intermittent tracking instability. For AR-TABLE where the sequences are significantly longer, our method attains the best average ATE of 5.94 cm, outperforming SplaTAM with ATE of 7.75\,cm and substantially improving over MonoGS. Per-sequence, SplaTAM is competitive on T3 and T5 but fails on several runs due to due to abrupt tracking jumps that trigger uncontrolled Gaussian insertion, while our method remains stable across all sequences; MonoGS attains low ATE on some sequences, but exhibits tracking failure with long sequences. This highlights a favorable accuracy-robustness trade-off in synthetic and real-world conditions.
\begin{table}[t]
	\centering
	\caption{The ATE of the estimated camera poses (cm) for five baseline methods and our VBGS-SLAM across different sequences of the AR-TABLE dataset.}
	\begin{tabular}{c|ccccccc|c}
		\toprule
		\textbf{Method} & \textbf{T1} & \textbf{T2} & \textbf{T3} & \textbf{T4} & \textbf{T5} & \textbf{T6} & \textbf{T7} & \textbf{Avg.}\\
		\midrule
		MonoGS & \textbf{3.42} & 5.03 & - & \textbf{2.43} & \textbf{2.01} & \textbf{6.87} & 26.06 & 7.64\\
        SplaTAM & 18.50 & 6.14 & \textbf{3.13} & - & \underline{3.22}& - & -  & 7.75\\
    \midrule
    \textbf{Ours} & 4.56 & \underline{4.22} & 7.57 & \underline{2.61} & 4.85 & 9.11 & \textbf{8.70} & \underline{5.94}\\
    \textbf{Ours} w/ IMU & \underline{4.13} & \textbf{3.95} & \underline{6.34} & 2.91 & 4.52 & \underline{8.97} & \underline{9.14} & \textbf{5.71}\\
		\bottomrule
	\end{tabular}
	\label{tab:artable_tracking}
    \vspace*{-10pt}
\end{table}
\vspace{-10pt}
\subsection{3D Reconstruction Performance}
\begin{figure*}[t]
    \centering
    \includegraphics[width=0.24\textwidth]{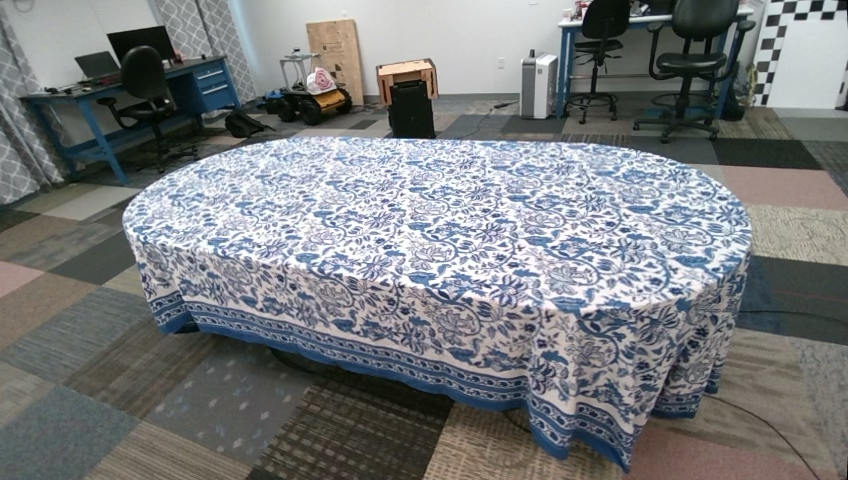}
    \includegraphics[width=0.24\textwidth]{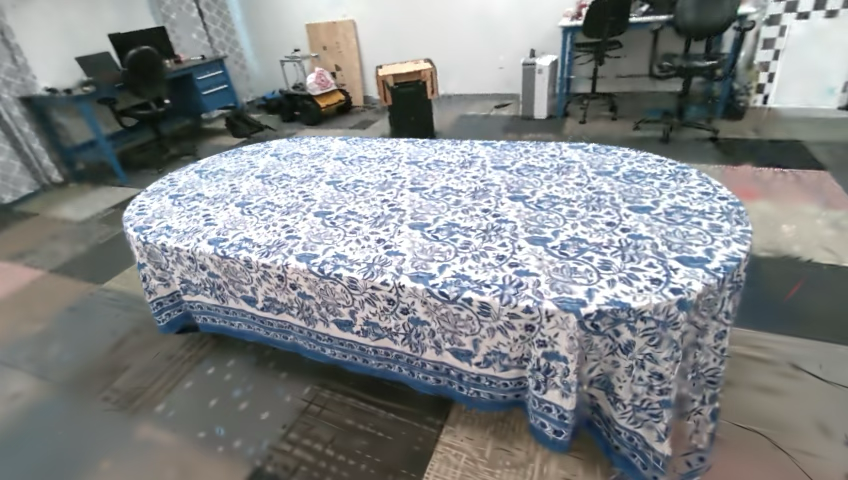}
    \includegraphics[width=0.24\textwidth]{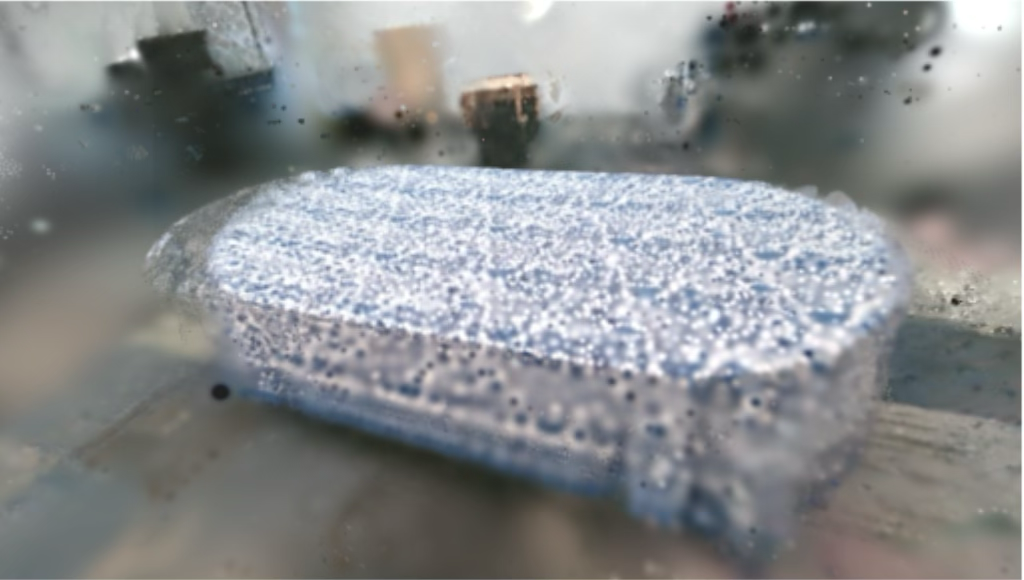}
    \includegraphics[width=0.24\textwidth]{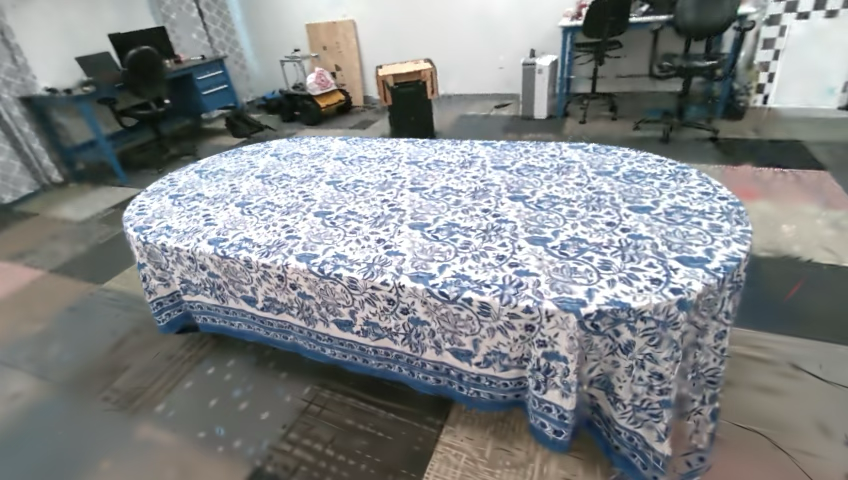}\\
    \vspace{0.1cm}
    \includegraphics[width=0.24\textwidth]{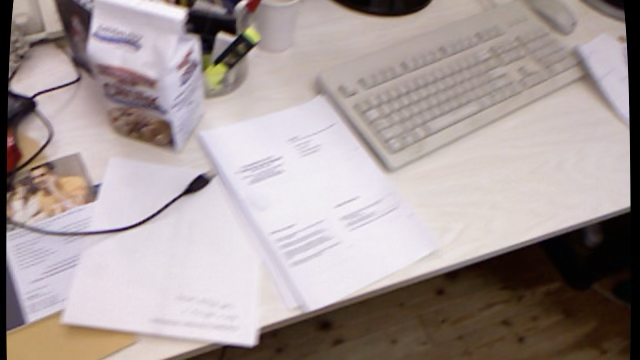}
    \includegraphics[width=0.24\textwidth]{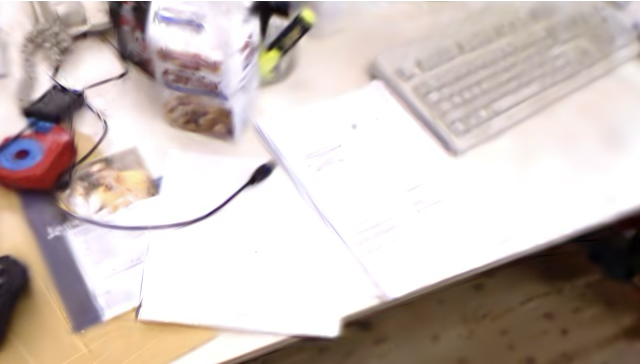}
    \includegraphics[width=0.24\textwidth]{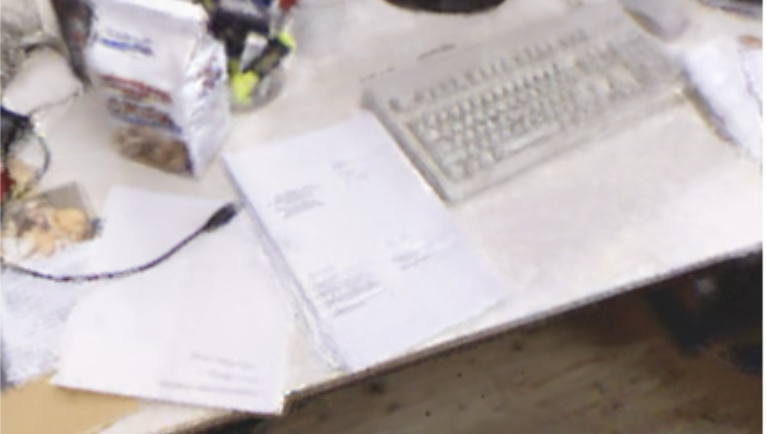}
    \includegraphics[width=0.24\textwidth]{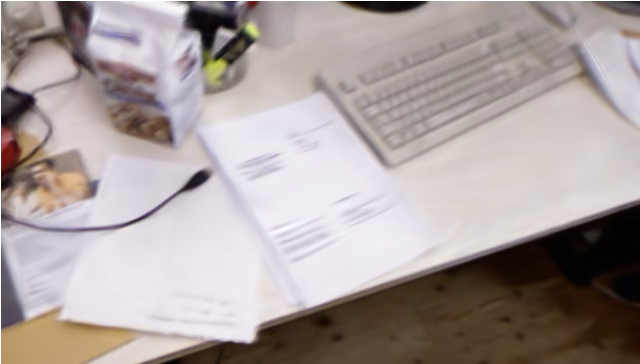}
	\caption{Qualitative Comparison of Rendering image to ground truth. The top row displays results on AR-TABLE dataset, while the second row showcases the render results on TUM-RGBD dataset. Columns correspond to rendering image of ground-truth, MonoGS, SplaTAM, and VBGS-SLAM (ours), respectively.}
	\label{fig:render_comp}
    \vspace*{-15pt}
\end{figure*}
\label{sec:recon_performance}
The rendering quality and efficiency are summarized on Replica, TUM-RGBD and AR-TABLE datasets in Table~\ref{tab:replica_rendering_performance}, ~\ref{tab:tum_rendering_performance}, ~\ref{tab:artable_rendering_performance}. On Replica, our approach yields high-fidelity novel view synthesis, achieving a PSNR of 37.94 dB, an SSIM of 0.95, and an LPIPS of 0.097, which is highly competitive with the leading methods for rendering, MonoGS, and significantly surpasses the fidelity of earlier systems like NICE-SLAM. We render at 0.465 FPS, slightly above MonoGS at 0.445 FPS, and well above SplaTAM at 0.115 FPS. On TUM-RGBD, we obtain the highest PSNR of 23.46 dB, second-best LPIPS of 0.31, with a second-best SSIM of 0.74; importantly, our renderer achieves 1.89 FPS, exceeding all methods except for Vox-Fusion. On AR-TABLE dataset, our method achieves a competitive PSNR of 35.24 dB and LPIPs of 0.079 competitive with MonoGS 37.79 dB, 0.077, and far above SplaTAM 18.15 dB, 0.348 -- while running at 1.374 FPS with the lowest GPU usage, roughly 2x faster than MonoGS and 5x faster than SplaTAM. This performance demonstrates our superior efficiency and prospects in real-time applications.
\begin{table}[h!]
\centering
\caption{Quantitative View Rendering Performance on Replica Dataset}
\label{tab:replica_rendering_performance}
\begin{tabular}{lcccc}
\toprule
\textbf{Method} & \textbf{PSNR[db]$\uparrow$} & \textbf{SSIM$\uparrow$} & \textbf{LPIPS$\downarrow$} & \textbf{FPS$\uparrow$} \\
\midrule
NICE-SLAM  & 24.42 & 0.809 & 0.233 & \underline{0.54} \\
Vox-Fusion  & 24.41 & 0.801 & 0.236 & \textbf{2.17} \\
Point-SLAM & 35.17 & \textbf{0.975} & 0.124 & 1.33 \\
SplaTAM & 34.11 & \underline{0.97} & 0.10 & 0.115 \\
MonoGS & \underline{37.79} & 0.96 & \textbf{0.077} & 0.445 \\
\midrule
\textbf{Ours} & \textbf{37.94} & 0.95 & \underline{0.097} & 0.465 \\
\bottomrule
\end{tabular}
\vspace*{-8pt}
\end{table}

\begin{table}[h!]
\centering
\caption{Quantitative View Rendering Performance on TUM-RGBD Dataset}
\label{tab:tum_rendering_performance}
\begin{tabular}{lcccc}
\toprule
\textbf{Method} & \textbf{PSNR[db]$\uparrow$} & \textbf{SSIM$\uparrow$} & \textbf{LPIPS$\downarrow$} & \textbf{FPS $\uparrow$} \\
\midrule
NICE-SLAM & 13.59 & 0.54 & 0.49 & 0.49 \\
Vox-Fusion & 15.54 & 0.63 & 0.50 & \textbf{2.17} \\
Point-SLAM & 15.63 & 0.66 & 0.53 & 0.22 \\
SplaTAM  & \underline{20.44} & \textbf{0.83} & \textbf{0.29} & 0.69 \\
MonoGS & 18.24 & 0.63 & 0.43 & 0.445 \\
\midrule
\textbf{Ours} & \textbf{23.46} & \underline{0.74} & \underline{0.31} & \underline{1.89} \\
\bottomrule
\end{tabular}
\vspace*{-15pt}
\end{table}

Fig.~\ref{fig:render_comp} compares ground-truth images with renderings from MonoGS, SplaTAM, and our method on AR-Table (top) and TUM-RGBD (bottom). On AR-Table, MonoGS preserves high-frequency detail but exhibits halos and minor “floaters’’ around depth discontinuities (e.g., the table boundary), whereas SplaTAM tends to over-smooth and inflate splats, washing out the cloth pattern and introducing blotchy artifacts. Our reconstruction retains the periodic texture of the tablecloth and delivers crisper, more coherent edges with fewer background smears. On TUM-RGBD where motion blur and lumination are pronounced, MonoGS remains sharp yet shows texture stretching/ghosting around fine structures, while SplaTAM again appears overly smoothed with fused edges. In contrast, our method better delineates thin structures and planar boundaries and suppresses speckle near discontinuities. Across both datasets, the renderings from our approach present fewer artifacts at depth edges, higher texture contrast, and improved perceptual coherence, aligning with the quantitative PSNR/LPIPS advantages reported in above.

\subsection{Ablation on IMU Propagation}
\label{sec:ablation_study}
We ablated the effect of IMU propagation on AR-Table by comparing VBGS-SLAM variants with and without IMU propagation and reported the tracking and rendering results in Table~\ref{tab:artable_tracking} and~\ref{tab:artable_rendering_performance}. Tracking accuracy changes only modestly: the average ATE improves from 5.94 cm without IMU to 5.71 cm with IMU, corresponding to a small gain of 0.23\,cm. Rendering quality shows similar minor differences, with the IMU-enabled variant achieving a PSNR of 36.75\,dB compared to 35.24\,dB without IMU, LPIPS of 0.075 versus 0.079, and identical SSIM of 0.78. Training throughput remains nearly unchanged at 1.374\,FPS, and GPU utilization is effectively identical. 

These results indicate that the proposed RGB-D variational pipeline already provides stable and accurate localization and reconstruction without relying on inertial propagation. In contrast to other GS-based SLAM systems that depend on ICP-stype geometric alignment or auxiliary motion cues for pose tracking, VBGS-SLAM achieves competitive performance through probabilistic pose inference. IMU propagation adds a slight boost in accuracy and perceptual quality without incurring additional runtime cost, further demonstrating the robustness of the proposed Bayesian pose estimation framework.

\setlength{\tabcolsep}{3pt}
\begin{table}[t]
\centering
\caption{Quantitative View Rendering Performance on AR-Table Dataset }
\label{tab:artable_rendering_performance}
\begin{tabular}{lccccc}
\toprule
\textbf{Method} & \textbf{PSNR[db]$\uparrow$} & \textbf{SSIM$\uparrow$} & \textbf{LPIPS$\downarrow$} & \textbf{FPS$\uparrow$}  & \textbf{GPU Uasage$\downarrow$}\\
\midrule
SplaTAM \cite{keetha2024splatam} & 18.15 & 0.69&  0.348 & 0.229 & 23.84\\
MonoGS \cite{matsuki2024gaussian} & \textbf{37.79} & \textbf{0.96} & \underline{0.077} & 0.445 & 20.86 \\
\midrule
\textbf{Ours} & 35.24 & 0.78 & 0.079 & \textbf{1.374} & \textbf{18.57} \\
\textbf{Ours} w/ IMU & \underline{36.75} & \underline{0.78} & \textbf{0.075} & \underline{1.373} & \underline{18.97} \\
\bottomrule
\end{tabular}
\vspace{-15pt}
\end{table}

\section{CONCLUSION}
We presented VBGS-SLAM, a fully probabilistic dense RGB-D SLAM framework that adapts Gaussian Splatting training with variational Bayesian inference. 
Our method achieves state-of-the-art accuracy and runtime on Replica, TUM-RGBD, and AR-TABLE, with robustness in early-stage tracking and reduced failure cases. The combination of closed-form update rules derived from sufficient statistics allows us to update effective observations without dense renderings. For future work, extending the model to dynamic scenes with motion priors and to multi-sensor settings would broaden the application.

\bibliography{references}
\bibliographystyle{IEEEtran}

\end{document}